\documentclass[runningheads]{llncs}
\usepackage{graphicx}
\usepackage{amsmath,amssymb} % define this before the line numbering.
\usepackage{color}

\usepackage{xcolor}
\usepackage{booktabs}
\usepackage{multirow}
\usepackage[hidelinks]{hyperref}

\usepackage[accsupp]{axessibility}  % Improves PDF readability for those with disabilities.

\newcommand{\shapelegend}{\includegraphics[height=2ex]{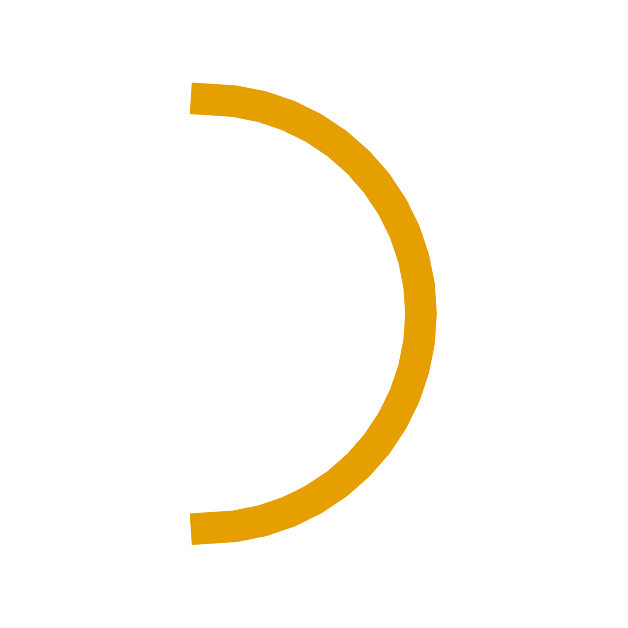}\includegraphics[height=2ex]{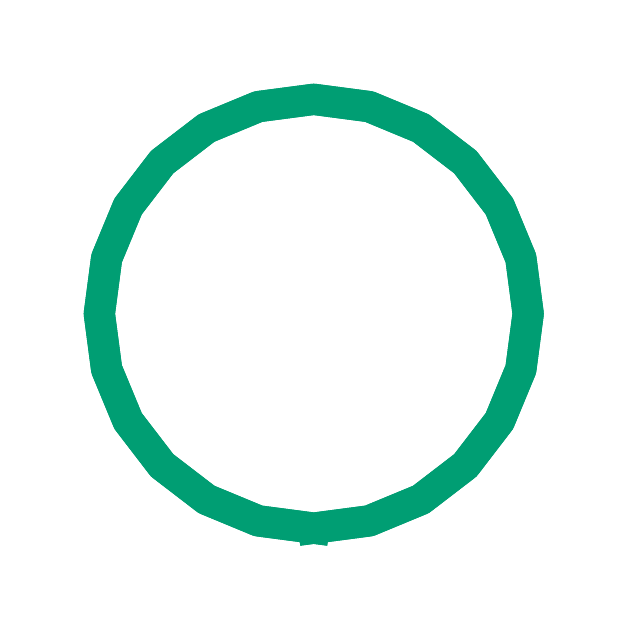}\includegraphics[height=2ex]{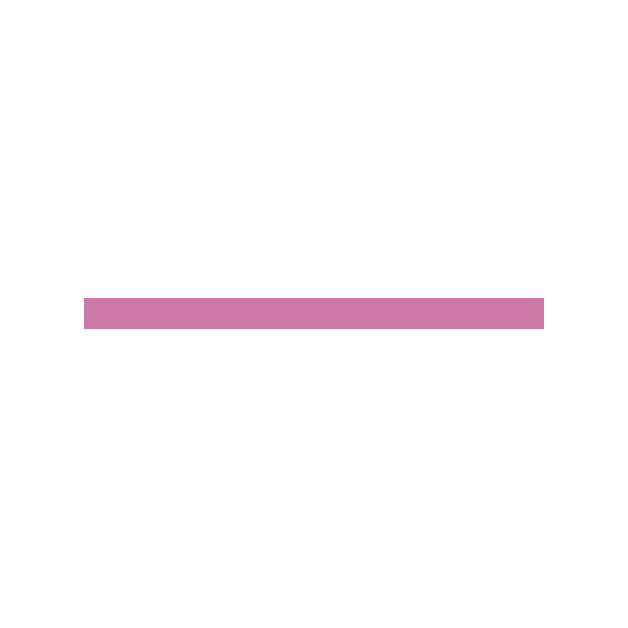}\includegraphics[height=2ex]{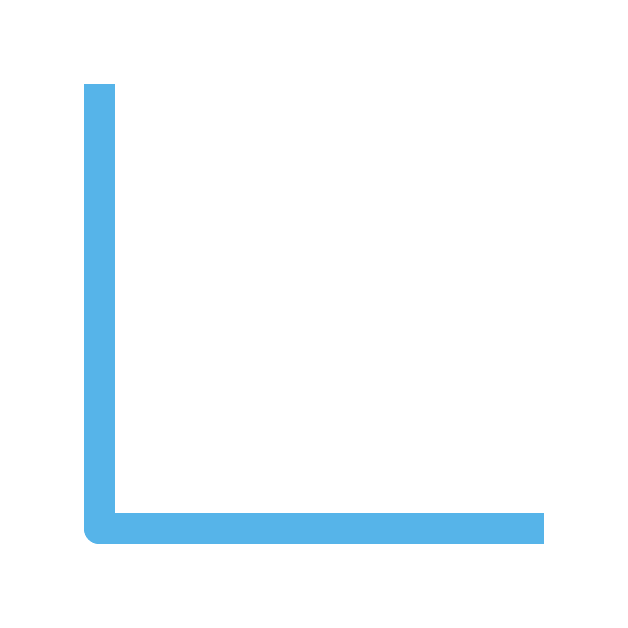}\includegraphics[height=2ex]{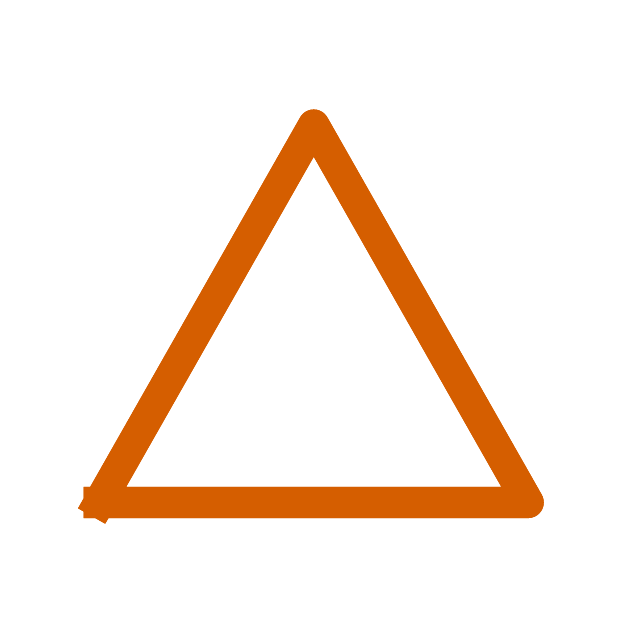}\includegraphics[height=2ex]{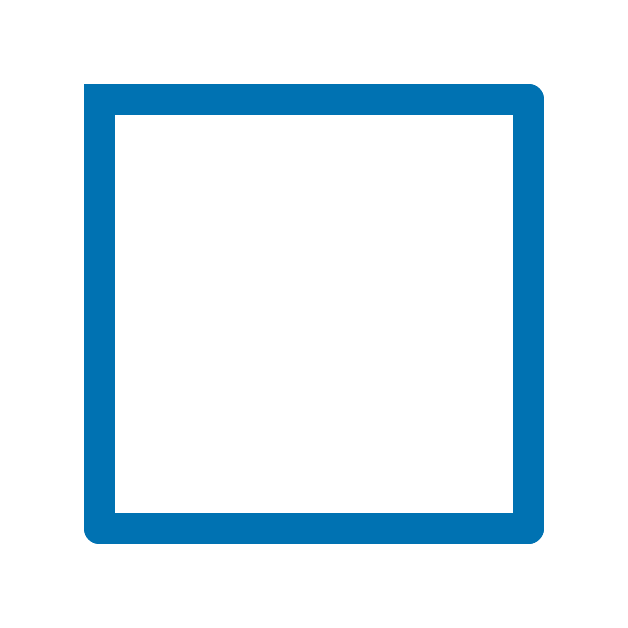}\includegraphics[height=2ex]{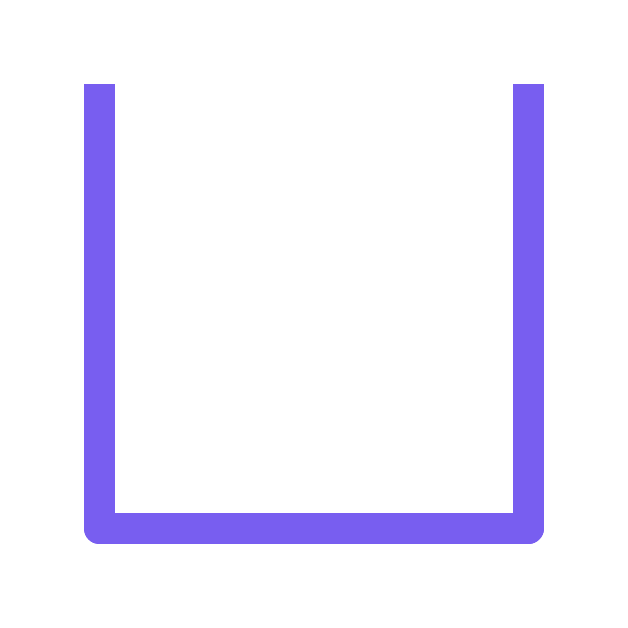}}

\begin{document}

\pagestyle{headings}
\mainmatter
\title{Abstracting Sketches through Simple Primitives} % Replace with your title

\titlerunning{Abstracting Sketches through Simple Primitives}

\author{Stephan Alaniz\inst{1,2} \and Massimiliano Mancini\inst{1}\and Anjan Dutta\inst{3} \and \\Diego Marcos\inst{4} \and Zeynep Akata\inst{1,2,5}\thanks{A. Dutta is with the Institute for People-Centred AI at the University of Surrey. S. Alaniz, M. Mancini and Z. Akata are with the Cluster of Excellence Machine Learning at the University of Tübingen.}}
\authorrunning{S. Alaniz et al.}
\institute{
University of T\"ubingen, \and
MPI for Informatics, \and University of Surrey,\\ \and  Wageningen University, \and MPI for Intelligent Systems
\\}
%******************
\maketitle

%%%%%%%%% ABSTRACT
\begin{abstract}
Humans show high-level of abstraction capabilities in games that require quickly communicating object information. They decompose the message content into multiple parts and communicate them in an interpretable protocol. Toward equipping machines with such capabilities, we propose the Primitive-based Sketch Abstraction task where the goal is to represent sketches using a fixed set of drawing primitives under the influence of a budget. To solve this task, our Primitive-Matching Network (PMN), learns interpretable abstractions of a sketch in a self supervised manner. Specifically, PMN maps each stroke of a sketch to its most similar primitive in a given set, predicting an affine transformation that 
aligns the selected primitive to the target stroke. We learn this stroke-to-primitive mapping end-to-end with a distance-transform loss that is minimal when the original sketch is precisely reconstructed with the predicted primitives. 
Our PMN abstraction empirically achieves the highest performance on sketch recognition and sketch-based image retrieval given a communication budget, while at the same time being highly interpretable. This opens up new possibilities for sketch analysis, such as comparing sketches by extracting the most relevant primitives that define an object category. Code is available at \url{https://github.com/ExplainableML/sketch-primitives}.
\end{abstract}

%%%%%%%%% BODY TEXT
\section{Introduction}
\label{sec:intro}
Consider the game \textit{Pictionary}\footnote{https://en.wikipedia.org/wiki/Pictionary},
where one player picks an object, e.g. a face, and draws the object in an iterative manner, e.g. using a large circle for the head, small lines for eyes and an arc for the mouth, until the other players guess the object correctly. 
The goal is to represent an object by 
decomposing it into parts that characterize this object using as few parts as possible such that another player can recognize it as fast as possible. 
The inherent human ability~\cite{finke1988explorations} that makes playing this game with multiple players possible is the ability to identify the most distinctive parts of the object and ground them into an interpretable communication protocol for the other players. In other words, humans are capable of a high level of abstraction when thinking about, recognizing and describing objects to other.

\begin{figure}[t]
    \centering
    \includegraphics[width=\linewidth]{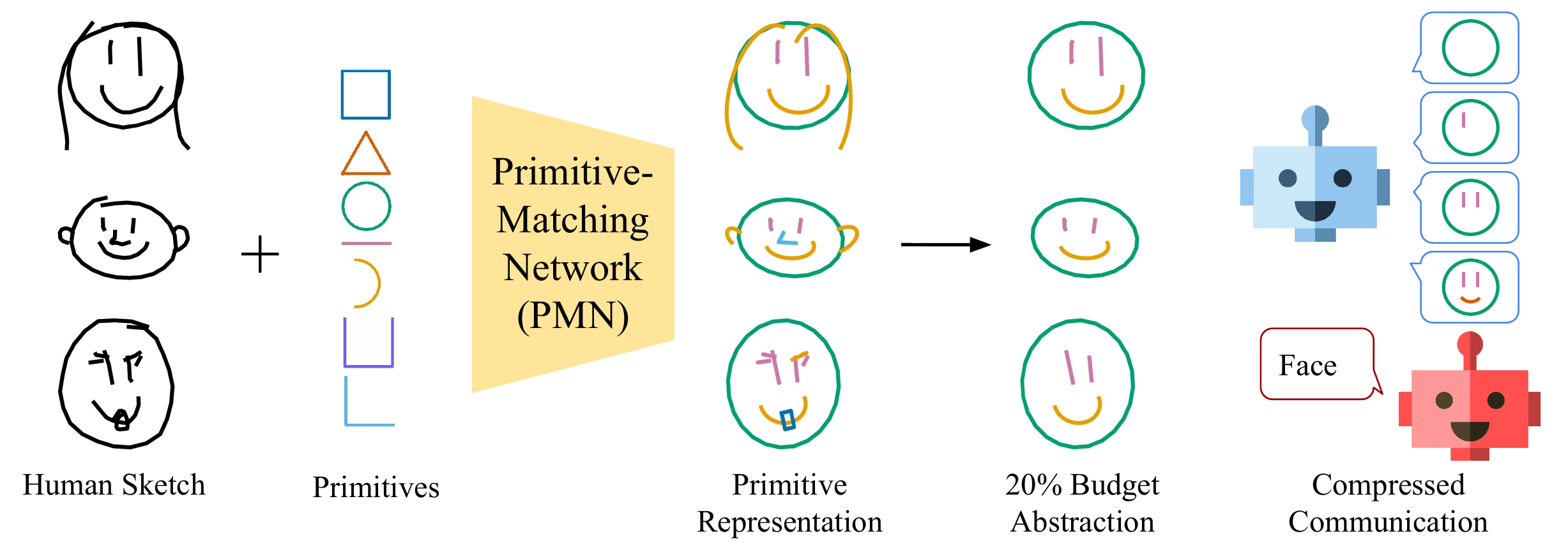}\vspace{-10pt}
    \caption{\textbf{Primitive-based Sketch Abstraction Task}. Our Primitive-Matching Network (PMN) takes human sketches and replaces their strokes with simple shapes from a set of 7 drawing primitives to create an abstract representation of the sketch. We can further compress the sketch by sub-selecting primitive strokes to meet a desired information budget. When communicating a sketch with a limited budget, our sketch abstractions retain the original semantics to perform well on downstream tasks.}\vspace{-10pt} 
    \label{fig:teaser}
\end{figure}

Inspired by this observation, we propose Primitive-based Sketch Abstraction as a new representation learning task, where the goal is to represent free-form drawings, i.e. sketches, by means of a fixed set of simple primitives. 
Sketches are an excellent tool for this task as they capture the essential parts of an object while removing the potentially adversarial texture and color information. However, humans have different drawing styles and skills, influenced by their upbringing and culture \cite{segall1966influence}. This causes different participants to draw the same instance of a real object in different ways (e.g. see Fig.~\ref{fig:teaser}, left).
We argue, however, that there exists a fundamental representation of each object class.
As demonstrated in \cite{finke1988explorations,roskos1993imagery,barquero1999imagery} when a participant draws an object of their imagination using a fixed dictionary of shapes providing a heavily abstracted representation of the object, another participant still guesses the object correctly. 

To solve the Primitive-based Sketch Abstraction task, we propose a self-supervised deep model, i.e. Primitive-Matching Network (PMN), to learn interpretable abstractions of a given object illustration without requiring any ground-truth abstraction. Differently from standard sketch-abstraction~\cite{MuhammadYSXH18,MuhammadYHXS19}, which selects subsets of the original strokes, {our model grounds them to a predefined vocabulary of primitives with a budget}, 
see Fig.~\ref{fig:teaser}. 
This way of representing sketches has two main advantages. First, it reduces the memory footprint of the sketch representation, allowing to communicate sketches by their constituent primitives rather than stroke coordinates. Second, it increases the interpretability of the sketch itself, making it much easier to compare and contrast sketches, e.g. a \textit{human face} is composed of a big circle for the head, two small lines for the eyes and one arc for the mouth whereas a \textit{cat face} is similar to a \textit{human face} but has triangles on top of the head for its ears.

Our PMN model replaces each stroke of a sketch with {a single} drawing primitive. 
This is achieved by mapping each stroke to its most similar primitive in a given set, and predicting an affine transformation that 
aligns the selected primitive to the target stroke. We train PMN by comparing the distance-transform of target strokes and their primitive-based version. At test time, given a sketch, we can efficiently choose a set of primitives and their spatial transformations, 
such that the generated sketch is fully composed of primitive shapes while being as similar as possible to the original one. 
Experiments on  sketch recognition and fine-grained sketch-based image retrieval tasks, show that the PMN abstraction achieves 
the highest performance given a communication budget (i.e. number of bytes necessary to communicate the sketch). Moreover, we show how we can use our abstraction to compare sketches, extracting the most relevant primitives and patterns that define an object category. 

To summarize, our contributions are: i) we propose the task of Primitive-based Sketch Abstraction,  where the goal is to produce interpretable sketch representations by means of
predefined drawing primitives; ii) we propose the first method for this task, Primitive-Matching Network, which learns to match strokes to primitives 
using as supervision a reconstruction loss over their distance transforms; iii) we show that PMN provides reliable sketch representations, communicating more information with a lower budget when compared with standard sketch abstraction methods, and eases sketch analysis.

\section{Related works}
\vspace{-5pt}
\paragraph{Sketch Abstraction.}
The goal of sketch abstraction \cite{MuhammadYSXH18,berger2013style} is to simplify the original strokes (or segments) from sketches without altering their semantic meaning. Abstracting sketches allows to communicate their information more effectively and efficiently, highlighting the most important traits of a sketch without corrupting its content \cite{berger2013style}. This is used in many applications, ranging from sketch-based image retrieval from edge-maps, to controllable sketch synthesis at various abstraction levels. Previous approaches addressed this problem through reinforcement learning, learning to remove sketch parts while preserving some desired features (e.g. semantic category, attributes) \cite{MuhammadYSXH18,MuhammadYHXS19}. Differently from previous works, we do not abstract sketches by removing strokes, but we ground them to a set of drawing primitives. This allows us to not only simplify the sketch representation itself, but to easily perform comparisons and analyses across sketches in a more straight forward manner than with stroke-based abstraction methods.

\vspace{-5pt}
\paragraph{Sketch Applications.} 
The release of the TU-Berlin \cite{eitz2010sketch} and QuickDraw \cite{jongejan2016quickdraw} datasets attracted the attention of the research community towards sketch classification.  Early works addressed the task with maximum margin classifiers over hand-crafted features \cite{li2013sketch,schneider2014sketch}. 
Advent of large-scale sketch datasets led to the development of deep learning models 
for this task that even surpassed human performance \cite{yu2017sketchanet}. Recent approaches explored deep and hand-crafted features
\cite{jia2020deepfeat}, multi-graph transformers \cite{xu2021multitrans},
coarse-to-fine hierarchical features 
\cite{yang2021sketchaa}, and learned tokenization schemes \cite{ribeiro2020sketchformer}.

Another popular application of sketches 
is \emph{sketch-based image retrieval} (SBIR), where the goal is to match free-hand sketches with corresponding natural images, both at category~\cite{parui2014sim,tolias2017asymm} and at instance level~\cite{bhunia2021morephotos,pang2020mixmodal}. 
Existing approaches for this task bridge the domain gap between photos and sketches by means of two branch architectures focusing on each modality independently
\cite{cao2011edgel,cao2010mindfinder,fedirici2020mib}, and even applying attention-based objectives \cite{song2017spatsematt,pang2019generalfgsbir} or self-supervised ones \cite{pang2020mixmodal}. Recently, \cite{BhuniaYHXS20} proposed to perform retrieval online, while the human is drawing. 
\cite{saavedra2015sketch,saavedra2017rst} perform SBIR by matching keyshapes to patches of sketch and contour images for SBIR, e.g. through S-HELO \cite{saavedra2014sketch} descriptors.
In this work, we do not directly address sketch recognition and SBIR,
but we use them to quantitatively analyze the compression/quality 
of our abstract sketch representations.

\vspace{-5pt}
\paragraph{Reconstruction with primitives.} One way of simplifying a complicated shape is to build an approximation using simple primitives. This is a central aspect of how humans understand the environment~\cite{biederman1987recognition} and has been applied to vector-like bitmap images~\cite{hammond2011recognizing,wu2015offline,reddy2021im2vec}{, CAD sketches \cite{para2021sketchgen,ganin2021computer,seff2021vitruvion},} and 3D shape reconstruction from sketches~\cite{smirnov2019deep} or images~\cite{liu2018physical,zeng2020bundle}. Interestingly, also many lossy image compression methods represent an image as a combination of predefined primitives~\cite{wallace1992jpeg,taubman2012jpeg2000}.
One closely related work~\cite{wu2015offline} focuses on diagram-like sketches, using shape proposals and an SVM classifier to assign the best-matching primitive. \cite{shapewords} represents sketches and edge maps of real images through lines and arcs for sketch-based image retrieval. Differently from these approaches, we are not restricted to specific domains \cite{wu2015offline}, or primitives \cite{shapewords}. PMN is generic and can be applied to any sketch, and any set of drawing primitives. 

\section{Abstracting Sketches by Drawing Primitives}
\label{sec:method}
Given a sketch, our goal is to obtain 
an abstract representation by replacing its strokes with a set of drawing primitives (e.g. squares, circles, lines). Formally, we have a training set $\mathcal{T}=\{\mathbf{s}^k\}_{k=1}^K$ of sketches, where $\mathbf{s}^k\in \mathcal{S}$ is a sketch in the set of possible drawings $\mathcal{S}$. Following previous works \cite{MuhammadYSXH18}, we assume that each sketch $\mathbf{s}^k$ (henceforth $\mathbf{s}$ for readability) is composed of a set of strokes (i.e. $\mathbf{s}=\{s_1,\dots,s_n\}$),
 and that each stroke is defined as a sequence of two-dimensional points of length $m(s_i)$. 
Additionally, we assume to have a set $\mathcal{P}\subset \mathcal{S}$ of drawing primitives that we want to use to represent our sketches, where each primitive is also a sequence of points. Note that no constraint is imposed on the primitives composing $\mathcal{P}$. 
At test time, given a sketch $\mathbf{s}$, our goal is to re-draw 
each stroke $s\in\mathbf{s}$ with a primitive $p\in \mathcal{P}$. This requires two steps: first, we need to map each stroke ${s}_i$ to its closest primitive ${p}_i\in \mathcal{P}$. Second, we need to compute the {affine transform} parameters 
making the primitive ${p}_i$ better fit the original stroke $s_i$. In the following, we describe how we achieve these goals. 

\subsection{Learning to match strokes and primitives}
\label{sec:method-training}

\begin{figure}[t]
    \centering
    \includegraphics[width=1.0\linewidth, trim={0 0 0 0}, clip]{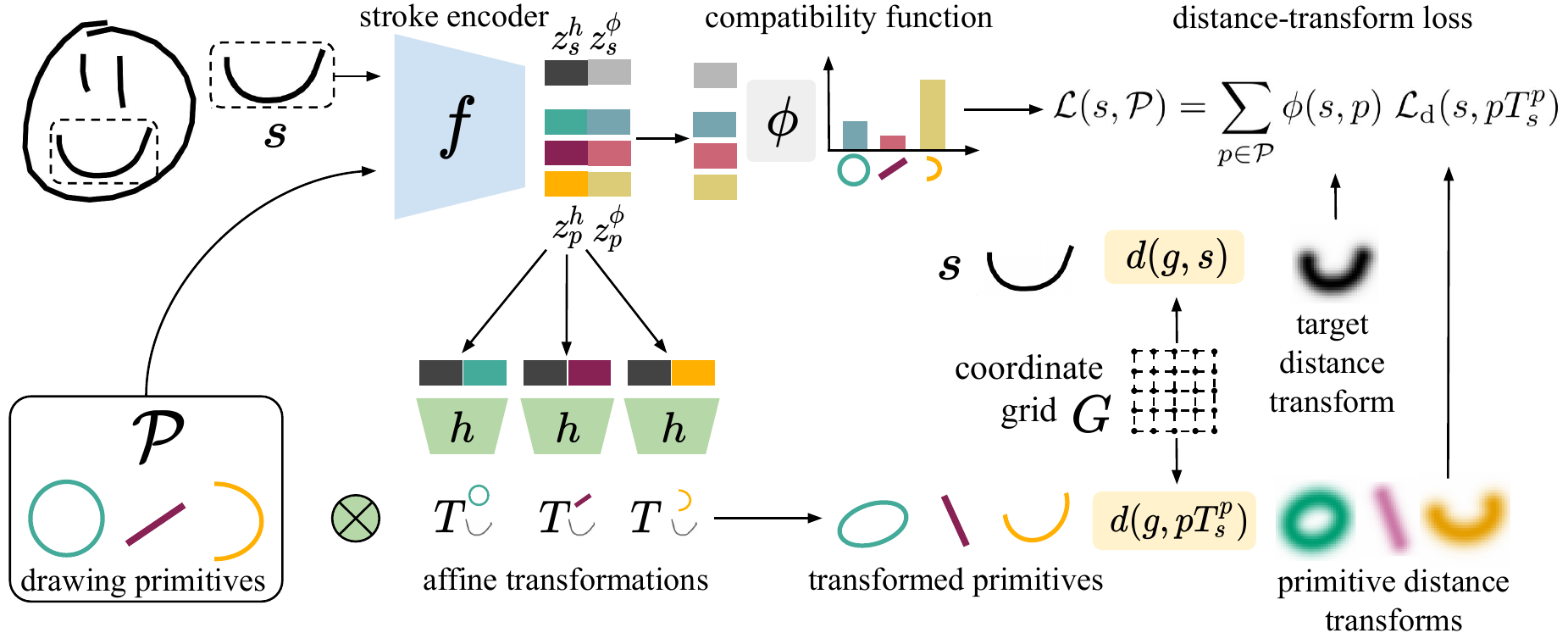}\vspace{-10pt}
    \caption{\textbf{PMN Model Architecture}. Given an input stroke (top left) and a set of primitives (bottom left), PMN encodes them into a shared embedding space using $f$. The embeddings are split in two parts, one for $h$ to compute the affine transformations aligning primitives to the target stroke, and one to compute the compatibility between primitives and the strokes with $\phi$. From a coordinate grid $G$, we compute a distance transform function of the stroke and the transformed primitives. We then use distance transforms and the compatibility scores to build the self-supervised objective of PMN.} \vspace{-10pt}
    \label{fig:model}
\end{figure}

There are two main challenges in matching an arbitrary stroke ${s}$ with a primitive $p\in\mathcal{P}$. First, we have no ground-truth pairs available, thus we have no direct information on which primitive $p$ is the most similar to the stroke $s$. Second, even if we find the best primitive, we need still to align it to the stroke. As a simple example, if we take two straight lines of equal length, a perfect match in case they are parallel, they result in a bad match if they are orthogonal to each other.  
We overcome these issues by i) applying an affine transformation to the primitives in $\mathcal{P}$ and ii) comparing the original strokes and the transformed primitives through their \emph{distance transform}. 

\paragraph{Aligning strokes and primitives.}
We need to transform a primitive in such a way that it better matches a given target stroke. To this end, we instantiate two functions, a stroke encoder $f:\mathcal{S}\rightarrow \mathbb{R}^d$, mapping a stroke (or primitive) to a d-dimensional embedding, and an alignment function 
$h:\mathbb{R}^d\times \mathbb{R}^d\rightarrow \text{Aff}(\mathbb{R}^2$), predicting the affine transformation 
that best aligns two strokes given their encoded representations. 
With $h$, we compute a transformation matrix $T^{p}_{s}$ as: 
\begin{equation}
    \label{eq:predicting-transformations}
    T^{p}_{s} = h(z_p, z_s)
\end{equation}
where $z_y = f(y)$ is the feature vector of the encoded sketch/primitive $y$, and $T^{p}_{s}$ the transformation aligning the primitive $p$ to the stroke $s$. 

\paragraph{Distance transform loss.}
Our goal is to find replacements for human strokes from a set of primitive shapes such that the visual difference is minimal. 
Given a stroke ${s}$, which is represented as a sequence of $m(s)$ connected points, i.e. $s=\{x_1,\dots, x_m\}$ and given a coordinate $g\in G$, with $G$ being a sampled coordinate grid, we can define the influence of the stroke at $g$ as:
\begin{equation}
\label{eq:distance-transform}
    d(g,{s}) = \max_{i \in \{1,\dots,m(s)-1\},\ r\in[0,1]}  \exp \big(-\gamma \; \lvert\lvert g-r\, x_i-(1-r)x_{i+1} \rvert\rvert^2 \big).
\end{equation}
Computing $d(g,s)$ for every coordinate in $G$ we obtain a distance map, also called \textit{distance transform} \cite{rosenfeld1968distance}. Note that in Eq.~\eqref{eq:distance-transform} we do not use directly the distance transform but its exponentially inverted version. This allows us to highlight the map on points closer to the stroke, with $\gamma$ acting as a smoothing factor. 
We can interpret this map 
as a visual rendering of the particular stroke, where the intensity of each pixel (coordinate) $g$ 
{decreases} with the distance of $g$ to the stroke itself. Considering a stroke $s$ and a primitive $p$, we can then define the distance transform loss as:
\begin{equation}
\label{eq:dt-loss}
\mathcal{L}_{\text{d}}(s, p{|h}) = \sum_{g\in G} \lvert\lvert d(g,s)- d(g,p{T^{p}_{s})}\rvert\rvert.
\end{equation}
With Eq.~\eqref{eq:dt-loss}, we are defining a reconstruction loss that sidesteps possible mismatches in the number of points contained in $s$ and $p$ as well as the need of matching points across the two strokes. For simplicity, we normalize the coordinates of each point in $s$ and $p$ to the range $[-1,1]$ before applying the loss and we consider $G$ as a set of linearly spaced coordinates in $[-1.5,1.5]$.

\paragraph{Exploiting stroke similarities.} Up to now we have discussed how we can align one primitive to a target stroke by means of the affine transformation computed by $h$ and how we can train $h$ by comparing distance transforms. However, during inference we want to replace $s$ with the best matching primitive selected from the set $\mathcal{P}$. With the current formulation, this could be done by replacing $s$ with the primitive $p\in \mathcal{P}$ for which the loss $\mathcal{L}_d(s,p|h)$ has the lowest value. 

While straightforward, this solution entails two issues. First, during inference we would need to compute the distance transform $d(g,p)$ for each $g\in G$ and $p\in \mathcal{P}$. Computing this map for each primitive is costly and would increase the inference time of the model. Second, if we do not consider how well a primitive $p$ matches a stroke $s$, 
we may have misleading training signals for $h$. To clarify, let us consider a simple example, where $s$ is a full circle and $p$ a simple straight line. In such case, the loss $\mathcal{L}_d(s,p|h)$ would be high even if $h$ predicts the best possible alignment. This means that the loss would be dominated by primitives that, such as $p$, cannot represent the stroke $s$, making $h$ focus on an ill-posed problem rather than on matching compatible primitive-stroke pairs. 

To address both issues, we inject the compatibility between a stroke and a primitive in the loss function. With this aim, we modify the stroke encoder as $f:\mathcal{S}\rightarrow \mathbb{R}^{2d}$ and, given an input $y$, we divide its embedding into two d-dimensional parts $z_y = [z_y^h, z_y^\phi] = f(s)$, where $z_y^h$ will be the part used to compute the alignment function through $h$ and $z_y^\phi$ will be used to compute the similarity between strokes/primitives. Given this embedding function, 
we calculate the 
relative similarity between a target stroke $s$ and a primitive $p$ as:
\begin{equation}
\label{eq:similarity}
    \phi(s,p) = \frac{\exp({\bar z_s^{\phi\intercal}} \bar z_p^\phi / \kappa)}{\sum_{q\in \mathcal{P}}\exp{({\bar z_s^{\phi\intercal}} \bar z_{q}^\phi} / \kappa)}
\end{equation}
where $\kappa$ is a temperature value, and $\bar{z}^\phi_y$ is the L2-normalized version of $z_y^\phi$.
Note that while $\phi$ needs to be invariant to the particular poses of $s$ and $p$ to score their compatibility, $h$ in Eq.~(\ref{eq:predicting-transformations}) needs to capture their pose to better align them. These conflicting objectives are what lead us to split the output of $f$ in two parts. With the compatibility scores, we can define our final loss as: 
\begin{equation}
\label{eq:full-loss}
    \mathcal{L}(s, \mathcal{P}|h,f) = \sum_{p \in \mathcal{P}} \phi(s,p) \; \mathcal{L}_{\text{d}}(s, p T^{p}_{s}).
\end{equation}
With this formulation, the lower the compatibility $\phi(s,p)$ between a primitive $p$ and the stroke $s$, the lower the weight of the distance transform loss between $p$ and $s$. 
Notably, the lowest value of $\mathcal{L}(s, \mathcal{P}|h,f)$ is achieved when i) the transformation matrices computed through $h$ align all primitives to the target stroke in the best way (w.r.t. the distance transforms), and ii) the primitives with the highest compatibility scores are the ones that better match the target stroke. Thus, minimizing $\mathcal{L}(s, \mathcal{P}|h,f)$ forces $h$ to output correct transformation matrices and $f$ to encode similar strokes close in the second half of the embedding space, fulfilling both our goals. We name the full model composed of $f$, $h$ and $\phi$ our \textit{Primitive Matching Network} (PMN). Fig.~\ref{fig:model} shows the PMN pipeline. 

\subsection{Replacing strokes with primitives}
\label{sec:method-inference}
After learning $f$ and $h$, we can replace strokes with primitives at test time. In particular, since computing the distance transform for each possible primitive is costly, we can directly use $f$ and $\phi$ to select the best matching primitive for a given stroke. Specifically, given a stroke $s$ of a sketch $\mathbf{s}$, we replace it by: 
\begin{equation}
\hat{p} = \arg\max_{p\in \mathcal{P}} \;\phi(s,p)
\end{equation}
where $\hat{p}$ is the best-matching primitive.
Given the primitive $\hat{p}$, we can now compute the corresponding alignment matrix as $T^{\hat{p}}_{s}$ from Eq.\eqref{eq:predicting-transformations}, 
and the abstracted sketch $\mathbf{\hat{s}}$ as:
\begin{equation}
    \label{eq:re-drawing}
   \mathbf{\hat{s}} = \{\hat{p_1}^\intercal T^{\hat{p_1}}_{s_1}, \cdots, \hat{p_n}^\intercal T^{\hat{p_n}}_{s_n},\}
\end{equation}
where $n=m(\mathbf{s})$ is the number of strokes in $\mathbf{s}$. We highlight that our formulation is agnostic to the number of strokes in a sketch, the shape and number of primitives in $\mathcal{P}$, and the number of points composing each stroke.

\section{Experiments}
In this section, we present our experimental results. We first discuss our experimental setting 
(Section~\ref{sec:exp-setup}) and show results on sketch classification (Section~\ref{sec:exp-cls}) and fine-grained sketch-based image retrieval (Section~\ref{sec:exp-sbir}) under a limited communication budget. Finally, we study the impact of 
the primitives 
(Section~\ref{sec:exp-ablation}) and show qualitative analysis on the abstract representations (Section~\ref{sec:exp-qualitative}).

\vspace{-15pt}
\subsection{Experimental setting} 
\begin{figure}[t]
    \centering
    \includegraphics[width=1.0\linewidth]{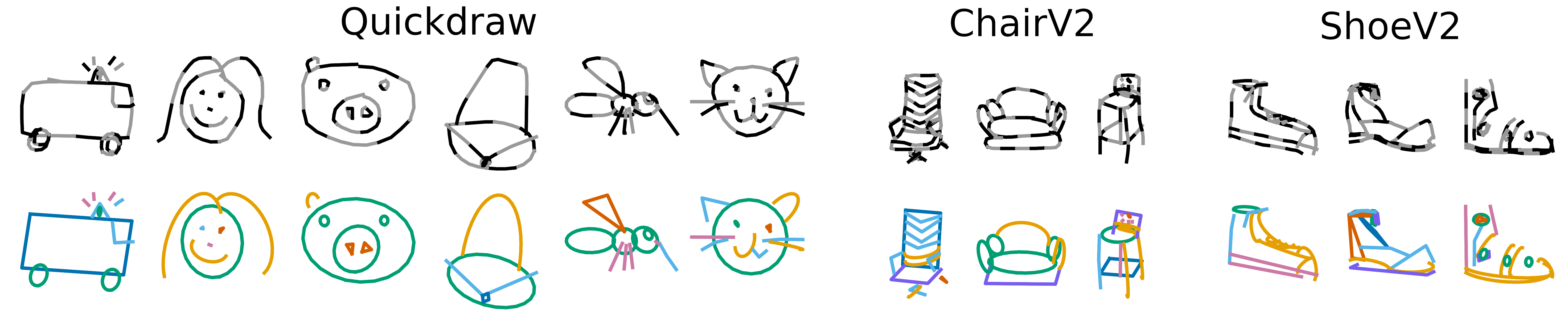}\vspace{-10pt}
    \caption{\textbf{Message content}. Human-drawn sketches (top) are split into messages of three points each (gray-coded). A primitive from the PMN representation (bottom) encodes more semantic information while requiring the same bytes for a single message.} \vspace{-10pt}
    \label{fig:sketches}
\end{figure}

\label{sec:exp-setup}

\vspace{-5pt}
\paragraph{Datasets and benchmarks.} Following previous works on sketch abstraction \cite{MuhammadYSXH18,MuhammadYHXS19} we test our model on sketch classification using Quickdraw~\cite{ha2018neural}, and on fine-grained sketch-based image retrieval (FG-SBIR) on ShoeV2~\cite{Yu2017} and ChairV2~\cite{Yu2017}. 

 For \textbf{Quickdraw}, we follow \cite{MuhammadYSXH18} and select,  630k sketches from nine semantic categories  (cat, chair, face, fire-truck, mosquito, owl, pig, purse and shoe). In this benchmark, we train a classifier on the original training sketches (details in the Supplementary), testing it on sketches abstracted using PMN or the competing methods given a specific budget (details below). We measure the performance as classification accuracy of the pretrained classifier given the abstracted inputs.

\textbf{ShoeV2} comprises 5982 training and 666 testing image-sketch pairs of various shoes. 
For this task, we train a Siamese network~\cite{song2017spatsematt} on the original training sketches with the same architecture of \cite{MuhammadYSXH18,BhuniaYHXS20}, 
replacing the standard triplet loss with a contrastive objective\footnote{We found the contrastive objective to stabilize and speed up the training without sacrificing retrieval accuracy.}.   
We measure the image-retrieval accuracy (top-10) of this network on test sketches abstracted using either PMN or one of competing methods, with the abstraction conditioned on a given budget. 

\textbf{ChairV2} contains 952 training and 323 testing pairs of chairs. For this dataset, we follow the FG-SBIR evaluation protocol described for ShoeV2.

\vspace{-15pt}
\paragraph{Implementation details.}
We train two neural networks $f$ and $h$ as described in Section~\ref{sec:method}. The stroke encoder $f$ is a 6-layer Transformer \cite{vaswani2017attention}, each with 8 self-attention heads. In all datasets, sketch data is represented as a list of points with their 2D coordinates and a binary label denoting whether the human is drawing or lifting the pen. We use the latter label to identify strokes.
We feed as input to $f$ the sequence of 2D-points of a stroke, together with an extra token used to obtain the final stroke embedding. 
We implement $h$ 
as a 3-layer MLP that takes the concatenated embedding $z_p^h$ and $z_s^h$ as input. We use 7 predefined primitives $\mathcal{P}$, as shown in Fig.~\ref{fig:teaser}, as they can represent
a wide variety of human strokes. 
We restrict $T^{p}_{s}$ to be a composite transformation of rotation, anisotropic scale, rotation in sequence, since we found it to be flexible enough to represent a wide variety of hand-drawn strokes. The translation is directly taken from the coordinates of $s$ and not predicted. The Supplementary contains more details about the transformation.
Hyperparameters $\gamma = 6$ and $\kappa = 0.2$ are  the same on all datasets and chosen by performing grid-search on a validation set.

\vspace{-5pt}
\paragraph{Budget computation.}
To quantify the level of abstraction of our primitive representation, we adopt a similar evaluation procedure as in \cite{MuhammadYHXS19}. Instead of measuring classification accuracy of full sketches, the evaluation is done at different subsets of the full sketch given a budget, amounting to different levels of sketch compression. Concretely, we test at budgets of 10\%, 20\% and 30\% of the original sketch's information content {to focus on the high compression regime}. 
To compute the budget, we follow the same procedure of \cite{MuhammadYSXH18}, considering a single message as made of three stroke points, i.e. three sets of 2D coordinates (see Fig.~\ref{fig:sketches}). 
Note that each message contains information equivalent to six floating points values and a categorical value indicating to which stroke the points belong. This is the same amount of information as a single primitive of our proposed model, defined as a 2D-translation, 2D-scale, 2D-rotation for its transformation and a categorical label indicating which of the 7 primitives is used. When evaluating the budget on human sketches, each message
corresponds to three points
of a hand-drawn stroke while, when using our abstract representation,
each message is a single primitive. Given a $N\%$ budget, we calculate the number of messages that can be communicated as the $N\%$ of the total 
messages forming the sketch. In Fig.~\ref{fig:sketches}, we illustrate what constitutes a message for human-drawn sketches and our primitive representations. 

\vspace{-5pt}
\paragraph{Compared methods.}
There are two ways in which a sketch can be abstracted. The first is by keeping the input representation untouched (i.e. original hand-draw strokes) but ranking the messages based on their importance for preserving the sketch content. Given a budget, we can then select only the most important subset of the messages. 
This is the approach of standard sketch abstraction methods \cite{MuhammadYSXH18,MuhammadYHXS19}. 
We categorize this strategy as \textit{Selection}-based abstraction. 

The second strategy is orthogonal and simplifies the sketch by grounding strokes to shapes in a fixed vocabulary, as in our PMN. This strategy does not define any ranking for the messages, but achieves abstraction by changing the stroke itself. We categorize this strategy under the name \textit{Shape}-based abstraction. In the experiments, we consider both type of approaches.

\textbf{Selection-based.} For this category, we consider two state-of-the-art methods: \textit{Deep Sketch Abstraction} (DSA)~\cite{MuhammadYSXH18} and \textit{Goal-Driven Sequential-data Abstraction} (GDSA)~\cite{MuhammadYHXS19}. 
DSA and GDSA are reinforcement learning methods that learn to order messages based on the performance on a downstream task. Specifically, DSA models the importance of each stroke by means of a classification (retrieval) rank-based reward, encouraging the target class (photo instance) to be highly ranked at all communication steps. GDSA is a more general strategy, applicable to various type of data. It directly uses the accuracy on the downstream task as reward function for the reinforcement learning agent, enforcing that the performance is preserved when the number of messages increases.

\textbf{Shape-based.} 
Since PMN is the first approach, we did not find other competitors in the literature addressing the same abstraction problem. As additional baseline we consider Shape Words (SW)~\cite{shapewords}, proposed in the context of sketch-based image retrieval. SW uses an heuristic algorithm to split the original strokes into multiple parts, fitting either a line or an arc to each part through Least Squares. 
Since SW cannot use arbitrary primitives, we use the same set of the original paper, i.e. lines and arcs. When PMN and SW are applied alone, the message order is the same
on which the original strokes were drawn. 

\textbf{Shape+Selection-based.} Since the two type of approaches are orthogonal, it is interesting to test if they can benefit each other. For this purpose, we also test other two models, combining GDSA with SW and our PMN.

\begin{table}[t]
    \centering
    \begin{tabular*}{.8\columnwidth}{@{\extracolsep{\stretch{0.1}}}*{6}{c}@{}}
    \multicolumn{2}{l}{Abstraction method} & 
    \multicolumn{4}{c}{Budget (\%)}\\
        Type & Name & 10 & 20 & 30 & 100\\
        \midrule
        \multirow{2}{*}{\shortstack[l]{Selection}} 
         & DSA~\cite{MuhammadYSXH18} & 20.12 & 43.85 & 64.04 & \multirow{2}{*}{\textbf{97.20}}\\
         & GDSA~\cite{MuhammadYHXS19} &  26.88 & 51.65 & 71.60 & \\[1mm]
        \multirow{2}{*}{\shortstack[l]{Shape}} &SW~\cite{shapewords} 
          & 51.21 & 68.20  & 75.60 & 78.30\\
         & PMN &  67.08  & 83.69 & 89.15 & 91.78 \\[1mm]
        \multirow{2}{*}{\shortstack[l]{Selection\\+Shape}} &SW+GDSA
          & 62.70 & 74.87 & 77.61 & 78.30\\
         & PMN+GDSA& \textbf{77.22} & \textbf{87.79} & \textbf{90.23} & 91.78 \\\hline
    \end{tabular*}\vspace{5pt}
    \caption{Classification accuracy on Quickdraw at budgets of 10\%, 20\% and 30\% evaluated with a classifier trained on the original human-drawn sketches.}
    \vspace{-20pt}
    \label{tab:cls_acc}
\end{table}

\vspace{-10pt}
\subsection{Sketch classification}
\label{sec:exp-cls}
In Tab.~\ref{tab:cls_acc}, we report the classification accuracy for both our PMN and
{the competitors} on Quickdraw for budgets 10\%, 20\%, 30\%, and 100\% as reference. From the experiments, we can see that methods SW and PMN, based on shape abstraction, outperform by a margin DSA and GDSA, based on message selection. This is a direct consequence of using shapes as messages rather than original stroke parts, since the former can communicate much more semantic information in a single message. 
We see the largest gain at low budgets, e.g. at a 10\% budget, DSA achieves 20.12\% accuracy, and GDSA 26.88\%, whereas SW reaches 51.21\% and PMN obtains 67.08\%, outperforming the rest significantly. This shows how PMN is better than SW at preserving the content of the sketch. This is a consequence of the higher flexibility in terms of 1) shapes that PMN can use and 2) precision of the alignment procedure, guided by the distance transform loss rather than Least Squares on heuristically selected points.  
The trend is similar at 20\% and 30\% budgets, at which point PMN achieves an accuracy of 89.18\% against 75.60\% of SW  and 71.60\% of GDSA. 
Notably, abstracting strokes with PMN is not lossless and the data distribution is different from the classifier's training data such that the accuracy at 100\% of PMN (91.78\%) is lower than using human sketches (97.20\%). 
On the up side, this allows PMN to reach an accuracy close to the upper bound of the original sketches at already 30\% budget showing that PMN 
well retains the semantic of the original dataset. 

Finally, if we couple a selection-based method (GDSA) with a shape-based ones, we see a consistent improvement of performance, with an improvement of 10\% (77.22\% accuracy) at 10\% budget for PMN+GDSA over simple PMN. Despite the improvement, SW+GDSA achieves lower performance than PMN alone at every budget (e.g. 62.70\% at 10\%), showing again how the abstraction of PMN is more precise than SW one.

\begin{table}[t]
    \centering
    \begin{tabular*}{\columnwidth}{@{\extracolsep{\stretch{1}}}*{6}{c}@{}@{\extracolsep{\stretch{1}}}*{4}{c}@{}}
   \multicolumn{2}{l}{Abstraction method} &  \multicolumn{4}{c}{ShoeV2, Budget (\%)}& 
    \multicolumn{4}{c}{ChairV2, Budget (\%)}\\\cmidrule{3-6}\cmidrule{7-10}
    Type & Name & 10 & 20 & 30 & 100 &10& 20  & 30 & 100\\
        \midrule
        \multirow{2}{*}{\shortstack[l]{Selection}} & DSA~\cite{MuhammadYSXH18} & 10.96 & 18.32 & 26.88 & \multirow{2}{*}{\textbf{75.22}} & 16.72  & 31.58 & 45.20& \multirow{2}{*}{\textbf{86.99}} \\
         & GDSA~\cite{MuhammadYHXS19} & 14.86 & 21.32 & 31.08 & & 20.74 & 33.13 & 47.68&\\[1mm]
        \multirow{2}{*}{\shortstack[l]{Shape}} & SW~\cite{shapewords} & 15.47 & 25.53 & 29.13 & 29.82 & 28.79 & 45.82  & 48.92 & 51.27\\
        &PMN &  29.58  & 48.50 & 54.35 & 56.55 & 53.87 & 70.59 & 73.99 & 75.92\\[1mm]
        Selection& SW+GDSA & 19.96 & 28.97 & 29.27 & 29.82 & 35.60 & 47.98 & 50.77 & 51.27 \\
        +Shape&PMN+GDSA & \textbf{36.18} & \textbf{50.45} & \textbf{55.10} & 56.55 & \textbf{63.15} & \textbf{73.68} & \textbf{75.23} & 75.92 \\\hline
    \end{tabular*}\vspace{5pt}
    \caption{Fine-grained sketch-based image-retrieval (FG-SBIR) results (top-10 accuracy) on ShoeV2 and ChairV2 at budgets of 10\%, 20\% and 30\% evaluated with a retrieval network trained on the original sketch-image pairs.}
    \vspace{-20pt}
    \label{tab:sbir}
\end{table}

\subsection{Sketch-based image retrieval}
\label{sec:exp-sbir}
In Tab.~\ref{tab:sbir}, we show the results of our PMN abstractions and the competing methods in the fine-grained sketch-based image retrieval (FG-SBIR) task for the ShoeV2 (left) and ChairV2 (right) datasets.  Similarly to classification, we report the results at three different budgets: 10\%, 20\% and 30\%. 

FG-SBIR has different challenges from sketch classification, since the communicated representation should precisely capture the specific characteristics of an instance rather than the shared ones of object categories. Despite our PMN abstraction smooths the specific details of strokes when grounding them to drawing primitives, it still preserves the most recognizable characteristics of an instance given a specific budget. Overall, the results are consistent with the ones on sketch classification, with PMN achieving the best results in both datasets and for each level of abstraction. For instance, at 10\% budget, PMN achieves a retrieval accuracy of $29.58\%$ on ShoeV2 and $53.87\%$ on ChairV2, surpassing by a comfortable margin SW (i.e. $15.47\%$ on ShoeV2, $28.79\%$ on ChairV2) and selection-based models (e.g. GDSA, $14.86\%$ on ShoeV2,  $20.74\%$ on ChairV2).

As a direct consequence of the inherent challenges of this FG-SBIR (requiring more detailed information), we see that the higher the budget, the higher the the gap between PMN and the competitors. With $30\%$ budget, PMN achieves $54.35\%$ accuracy on ShoeV2 and $73.99\%$ on ChairV2 
while SW best result is $29.13\%$ on ShoeV2 and $49.92\%$ on ChairV2 and GDSA achieves $31.08\%$ on ShoeV2 and $47.68\%$ on ChairV2.
SW shows an opposite trend, with the performance gap with selection-based methods becoming smaller as the budget increases, performing lower than GSDA on ShoeV2 for a 30\% budget. These results highlight that PMN makes a more precise use of the available primitives, achieving the best trade-off between compression and distinctiveness of the sketch representation.

As expected, coupling PMN with GDSA leads to the best results overall (e.g. 36.18\% on ShoeV2 and 63.15\% on ChairV2 at 10\%), with the performance of PMN alone consistently surpassing the ones of SW+GDSA (e.g. 19.96\% on ShoeV2 and 35.60\% on ChairV2 at 10\%), highlighting that while selection and shape-based methods are complementary, it is fundamental that the
latter precisely reconstructs the input, something achieved by PMN and not by SW.

\begin{table}[t]
    \centering
    \begin{tabular*}{\columnwidth}{@{\extracolsep{\stretch{1}}}*{1}{l}@{}@{\extracolsep{\stretch{1}}}*{1}{c}@{}@{\extracolsep{\stretch{1}}}*{3}{c}@{}@{\extracolsep{\stretch{1}}}*{1}{c}@{}@{\extracolsep{\stretch{1}}}*{3}{c}@{}}
        & \multicolumn{4}{c}{Quickdraw (Classification)} & \multicolumn{4}{c}{ChairV2 (FG-SBIR)} \\\cmidrule{2-5}\cmidrule{6-9}
        \multirow{2}{*}{Primitives}&\multirow{2}{*}{Usage (\%)}& \multicolumn{3}{c}{Budget (\%)} & \multirow{2}{*}{Usage (\%)}& \multicolumn{3}{c}{Budget(\%)} \\
         & & 10 & 20 & 30 &  & 10 & 20 & 30\\
        \midrule
         \includegraphics[height=2ex]{imgs/primitives/arc.pdf} & 22.72 & 45.99 & 70.58 & 79.82 & 39.17 & 41.79 & 62.53 & 69.34\\
         + \includegraphics[height=2ex]{imgs/primitives/circle.pdf} & 20.97 & 62.48 & 79.68 & 86.15 & 10.63 & 42.72 & 63.77 & 69.65\\
         \hspace{2mm}+ \includegraphics[height=2ex]{imgs/primitives/line.pdf} & 19.60 & 62.04 & 79.92 & 86.63 & 15.89 & 43.03 & 64.08 & 69.65\\
         \hspace{4mm}+ \includegraphics[height=2ex]{imgs/primitives/corner.pdf} & 15.84 & 64.07 & 81.26 & 87.81 & 13.81 & 43.96 & 64.70 & 69.96\\
         \hspace{6mm}+ \includegraphics[height=2ex]{imgs/primitives/triangle.pdf} & 8.67 & 64.98 & 82.64 & 88.85 & 7.57 & 44.58 & 65.94 & 70.27\\
         \hspace{8mm}+ \includegraphics[height=2ex]{imgs/primitives/square.pdf} & 6.20 &  66.93 & 83.59 & 89.12 & 5.79 & 49.22 & 69.34 & 73.37\\
         \hspace{10mm}+ \includegraphics[height=2ex]{imgs/primitives/u.pdf} & 6.00 & 67.08 & 83.69 & 89.15 & 7.12 & 53.87 & 70.59 & 73.99\\\hline
    \end{tabular*}\vspace{5pt}
    \caption{Results of PMN with different 
    primitives on Quickdraw (acc.) and ChairV2 (top-10 acc.). Primitives added one at a time in order of usage in Quickdraw.}\vspace{-20pt}
    \label{tab:cls_acc_ablation}
\end{table}

\subsection{Ablation study}
\label{sec:exp-ablation}
In Tab.~\ref{tab:cls_acc_ablation}, we analyze the importance of the primitive shapes by evaluating the PMN model with different subsets of primitives for Quickdraw and ChairV2. We use the PMN model trained with all seven primitives 
and at test time only provide a subset of them. 
We start with the most commonly used shape, the arc, and add one primitive at a time in the order of their usage frequency in the Quickdraw dataset. While the arc alone does not provide enough flexibility in Quickdraw, with only two primitives, arc and circle, our PMN model achieves a higher classification accuracy than both GDSA and SW with 62.48\% (vs. 51.21\% in SW) at a 10\% budget and 86.15\% (vs. 75.60\% in SW) at a 30\% budget.
To put these results into perspective, SW is able to represent line, circle and arc shapes, so even without using lines the PMN model can better fit shapes and reconstruct sketches while retaining their semantics. This is particularly evident for ChairV2, where, even by using only arcs, PMN surpasses SW at all budgets (e.g. 41.79\% vs 28.79\% at 10\% budget).

As more shapes are added, there are diminishing returns in increasing classification accuracy for Quickdraw. However, every primitive contributes to the performance our model achieves. The triangle, square and U-shape stand out to provide a significant improvement despite their relatively low usage of 8.67\%, 6.20\% and 6.00\% respectively. Interestingly, on ChairV2 we see a more monotonic increase in the performance (e.g. from 44.58\% to 49.22\% when adding squares at 10\% budget). This is expected, since FG-SBIR requires a more precise reconstruction of the original sketch, thus having more primitives helps in better modeling the specificity of each stroke, improving the overall results. 

As a final note, while there are many options on which primitives to include, these results validate the choice of these seven primitives. Nonetheless, depending on the dataset and use case, other choices could be considered.

\begin{figure}[t]
    \centering
    \includegraphics[width=1.0\linewidth]{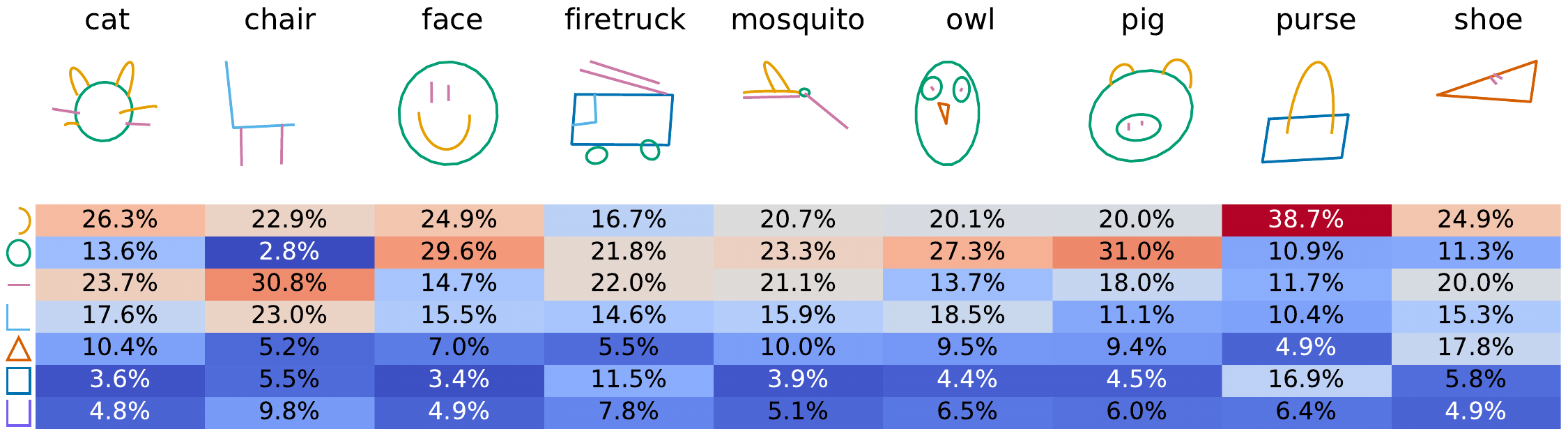}\vspace{-5pt}
    \caption{\textbf{Primitive analysis on Quickdraw}. Primitive representation of each Quickdraw class and the usage frequency of each primitive per class (in \%).} \vspace{-5pt}
    \label{fig:quickdraw-classes-qual}
\end{figure}

\subsection{Qualitative analysis}
\label{sec:exp-qualitative}
\vspace{-5pt}
\paragraph{How are objects represented through primitives?} An interesting aspect of PMN is that we can now compare strokes across different sketches, extracting
possible patterns. 
To show one possible application, in Fig.~\ref{fig:quickdraw-classes-qual}, we analyze the use of primitives 
when reconstructing Quickdraw classes. We show a representative abstracted sample of each class and 
the distribution of the primitives per class. 

When inspecting the primitive distributions, we observe that the most used primitives are arcs and circles. As shown in our ablation study (cf. Tab.~\ref{tab:cls_acc_ablation}), using these two primitives alone can already cover a lot of variation  
on human strokes. Common use cases for arcs include the ears in animals, smiles in faces and handles in purses. Circles most frequently represent heads in animals and faces and firetrucks' wheels. The body of the firetruck and the purse are often represented by rectangles. These correlations can be observed when
comparing the average distribution of primitives per class, e.g. more frequent use of line and corner in chairs or rectangle and arc in purses than in other classes.

Fig.~\ref{fig:quickdraw-classes-qual} also shows a limitation of our PMN model. PMN tries to match one primitive to each human stroke. However, when a stroke cannot be easily represented by a primitive, PMN may provide inaccurate representations. This is the case of the shoe class, where the main part of the shoe is usually drawn in a single stroke with a closed L-shape. In this case, PMN approximates this L-shape with a triangle (17.8\% of shoe primitives are triangles, more than in any other class) that, despite driving the semantic of the sketch, provides a less accurate abstraction. In the future it would be interesting to address such cases by either learning to split/merge strokes and their parts into simpler shapes or by learning the primitives $\mathcal{P}$ together with PMN.

\begin{figure}[t]
    \centering
    \begin{minipage}{.51\textwidth}
      \centering
      \includegraphics[width=\linewidth]{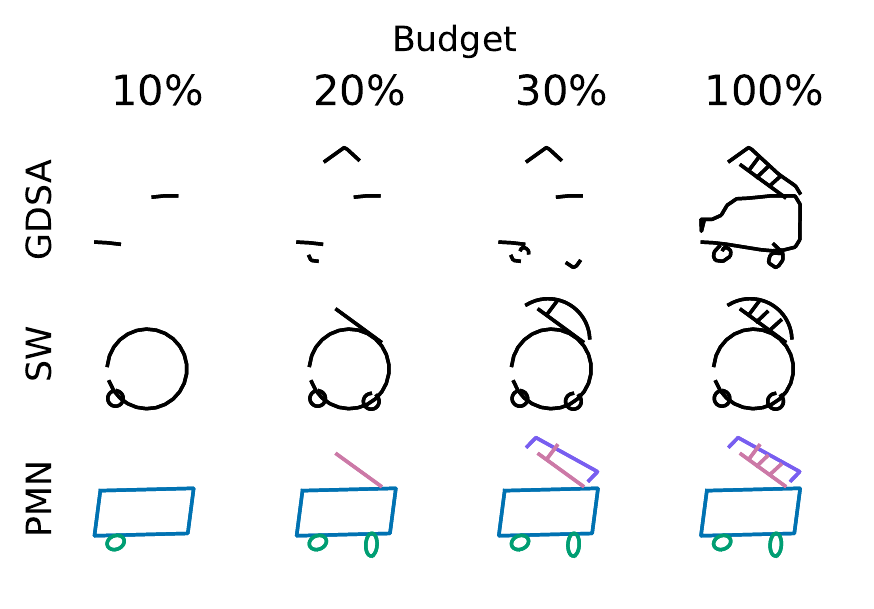}
      \includegraphics[width=\linewidth]{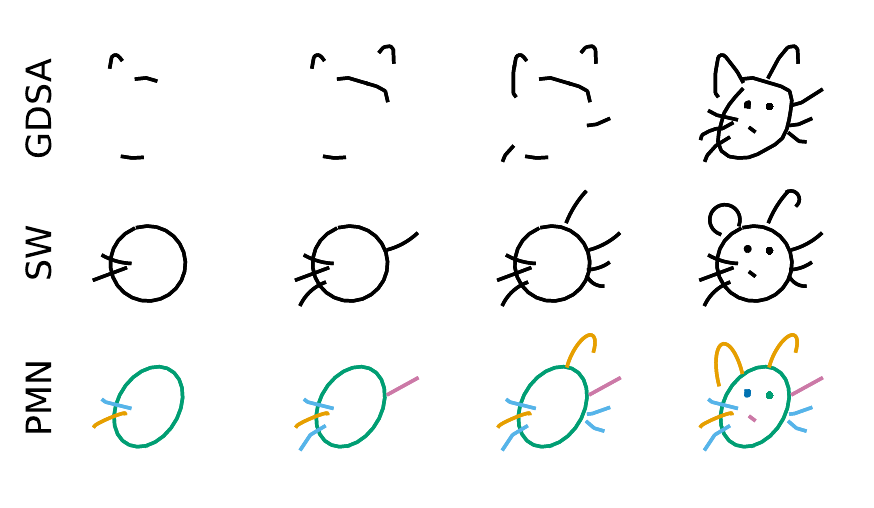}
    \end{minipage}
    \begin{minipage}{.48\textwidth}
      \centering
      \includegraphics[width=\linewidth]{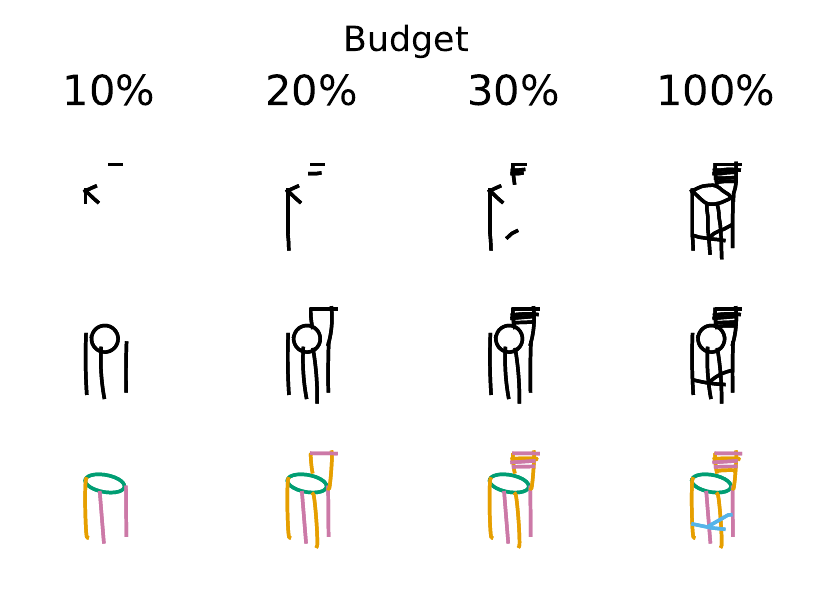}
      \includegraphics[width=\linewidth]{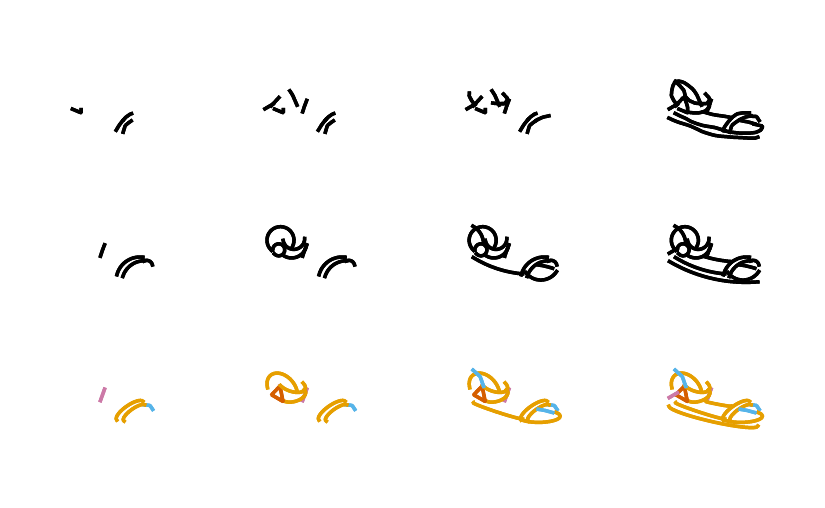}
    \end{minipage}
    \vspace{-10pt}
    \caption{\textbf{Qualitative example of sketches at different budgets}. We show example sketches of Quickdraw (left), ChairV2 (top right) and ShoeV2 (bottom right) at 10\%, 20\%, 30\% budgets when using GDSA, SW, PMN. Primitive color legend: \shapelegend.}
    \vspace{-10pt}
    \label{fig:classification-budget-qual}
\end{figure}

\vspace{-5pt}
\paragraph{Representations at different budgets.}
We inspect some example qualitative results of our model in Fig.~\ref{fig:classification-budget-qual}, showing sketch abstractions with varying compression budgets. We can see that partitioning the original strokes into three-point sub-strokes results in unrecognisable sketches even if GDSA is used to optimize the selection order. On the other hand, both SW and our PMN preserve the semantic much better given the same budget levels (e.g. shoe, bottom right). However, the additional flexibility allowed by PMN results in a much more faithful abstraction than SW, as exemplified by the body and ladder of the firefighting truck (top left), which are both represented by unnaturally rounded shapes by SW. Even when SW and PMN select the same shapes, PMN better aligns them to the original stroke, as can be seen 
from the circle used as seat of the chair (top right), or the arc used as left-ear of the cat (bottom left). This confirms the advantage of our self-supervised alignment objective w.r.t. the less flexible Least Squares solution of SW.  

\section{Conclusion}
Motivated by how humans abstract object representations in interpretable messages when playing communication games, in this paper we proposed a new representation learning task, Primitive-based Sketch Abstraction, where the goal is to represent a sketch with a given set of simple drawing primitives. To address this task we proposed a model, Primitive-Matching Network, that maps each stroke of a sketch to its closest drawing primitive and predicts the affine transformation to align them. We overcome the lack of annotation for stroke abstractions by developing a self-supervised objective 
using the distance transforms of the primitives and the target strokes. Experiments show that our model surpasses standard sketch abstraction methods on sketch classification and sketch-based image retrieval at a given budget. Differently from hand-drawn strokes, our PMN abstraction is highly interpretable and leads to new types of sketch analyses, comparing sketches by means of their primitives.

\vspace{-10pt}
\paragraph{\newline \textbf{Acknowledgments}.}
This work has been partially funded by the ERC (853489 - DEXIM), by the DFG (2064/1 – Project number 390727645), and as part of the Excellence Strategy of the German Federal and State Governments.
\bibliographystyle{splncs04}
\bibliography{egbib}

\begin{thebibliography}{10}
\providecommand{\url}[1]{\texttt{#1}}
\providecommand{\urlprefix}{URL }
\providecommand{\doi}[1]{https://doi.org/#1}

\bibitem{barquero1999imagery}
Barquero, B., Logie, R.: Imagery constraints on quantitative and qualitative
  aspects of mental synthesis. European Journal of Cognitive Psychology
  \textbf{11}(3) (1999)

\bibitem{berger2013style}
Berger, I., Shamir, A., Mahler, M., Carter, E., Hodgins, J.: Style and
  abstraction in portrait sketching. ACM Transactions on Graphics (TOG)
  \textbf{32}(4) (2013)

\bibitem{bhunia2021morephotos}
Bhunia, A.K., Chowdhury, P.N., Sain, A., Yang, Y., Xiang, T., Song, Y.Z.: More
  photos are all you need: Semi-supervised learning for fine-grained sketch
  based image retrieval. In: CVPR (2021)

\bibitem{BhuniaYHXS20}
Bhunia, A.K., Yang, Y., Hospedales, T.M., Xiang, T., Song, Y.: Sketch less for
  more: On-the-fly fine-grained sketch-based image retrieval. In: CVPR (2020)

\bibitem{biederman1987recognition}
Biederman, I.: Recognition-by-components: a theory of human image
  understanding. Psychological review  \textbf{94}(2) (1987)

\bibitem{cao2011edgel}
Cao, Y., Wang, C., Zhang, L., Zhang, L.: Edgel index for large-scale
  sketch-based image search. In: CVPR (2011)

\bibitem{cao2010mindfinder}
Cao, Y., Wang, H., Wang, C., Li, Z., Zhang, L., Zhang, L.: Mindfinder:
  interactive sketch-based image search on millions of images. In: ACM MM
  (2010)

\bibitem{eitz2010sketch}
Eitz, M., Hildebrand, K., Boubekeur, T., Alexa, M.: Sketch-based image
  retrieval: Benchmark and bag-of-features descriptors. IEEE transactions on
  visualization and computer graphics  \textbf{17}(11) (2010)

\bibitem{fedirici2020mib}
Federici, M., Dutta, A., Forré, P., Kushman, N., Akata, Z.: {Learning Robust
  Representations via Multi-View Information Bottleneck}. In: ICLR (2020)

\bibitem{finke1988explorations}
Finke, R.A., Slayton, K.: Explorations of creative visual synthesis in mental
  imagery. Memory \& cognition  \textbf{16}(3) (1988)

\bibitem{ganin2021computer}
Ganin, Y., Bartunov, S., Li, Y., Keller, E., Saliceti, S.: Computer-aided
  design as language. NeurIPS  (2021)

\bibitem{ha2018neural}
Ha, D., Eck, D.: A neural representation of sketch drawings. In: ICLR (2018)

\bibitem{hammond2011recognizing}
Hammond, T., Paulson, B.: Recognizing sketched multistroke primitives. ACM
  Transactions on Interactive Intelligent Systems (TiiS)  \textbf{1}(1) (2011)

\bibitem{hendrycks2016gelu}
Hendrycks, D., Gimpel, K.: Gaussian error linear units (gelus). arXiv preprint
  arXiv:1606.08415  (2016)

\bibitem{jia2020deepfeat}
Jia, Q., Fan, X., Yu, M., Liu, Y., Wang, D., Latecki, L.J.: Coupling deep
  textural and shape features for sketch recognition. In: ACM MM (2020)

\bibitem{jongejan2016quickdraw}
Jongejan, J., Rowley, H., Kawashima, T., Kim, J., Fox-Gieg, N.: {The Quick,
  Draw! - A.I. Experiment}. https://quickdraw.withgoogle.com  (2016)

\bibitem{li2013sketch}
Li, Y., Song, Y., Gong, S.: Sketch recognition by ensemble matching of
  structured features. In: BMVC (2013)

\bibitem{liu2018physical}
Liu, Z., Freeman, W.T., Tenenbaum, J.B., Wu, J.: Physical primitive
  decomposition. In: ECCV (2018)

\bibitem{MuhammadYHXS19}
Muhammad, U.R., Yang, Y., Hospedales, T.M., Xiang, T., Song, Y.: Goal-driven
  sequential data abstraction. In: ICCV (2019)

\bibitem{MuhammadYSXH18}
Muhammad, U.R., Yang, Y., Song, Y., Xiang, T., Hospedales, T.M.: Learning deep
  sketch abstraction. In: CVPR (2018)

\bibitem{pang2019generalfgsbir}
Pang, K., Li, K., Yang, Y., Zhang, H., Hospedales, T.M., Xiang, T., Song, Y.Z.:
  Generalising fine-grained sketch-based image retrieval. In: CVPR (2019)

\bibitem{pang2020mixmodal}
Pang, K., Yang, Y., Hospedales, T.M., Xiang, T., Song, Y.Z.: Solving
  mixed-modal jigsaw puzzle for fine-grained sketch-based image retrieval. In:
  CVPR (2020)

\bibitem{para2021sketchgen}
Para, W., Bhat, S., Guerrero, P., Kelly, T., Mitra, N., Guibas, L.J., Wonka,
  P.: Sketchgen: Generating constrained cad sketches. NeurIPS  (2021)

\bibitem{parui2014sim}
Parui, S., Mittal, A.: Similarity-invariant sketch-based image retrieval in
  large databases. In: ECCV (2014)

\bibitem{reddy2021im2vec}
Reddy, P., Gharbi, M., Lukac, M., Mitra, N.J.: Im2vec: Synthesizing vector
  graphics without vector supervision. In: CVPR (2021)

\bibitem{ribeiro2020sketchformer}
Ribeiro, L.S.F., Bui, T., Collomosse, J., Ponti, M.: Sketchformer:
  Transformer-based representation for sketched structure. In: Proceedings of
  the IEEE/CVF conference on computer vision and pattern recognition. pp.
  14153--14162 (2020)

\bibitem{rosenfeld1968distance}
Rosenfeld, A., Pfaltz, J.L.: Distance functions on digital pictures. Pattern
  recognition  \textbf{1}(1) (1968)

\bibitem{roskos1993imagery}
Roskos-Ewoldsen, B., Intons-Peterson, M.J., Anderson, R.E.: Imagery,
  creativity, and discovery: A cognitive perspective. Elsevier (1993)

\bibitem{saavedra2014sketch}
Saavedra, J.M.: Sketch based image retrieval using a soft computation of the
  histogram of edge local orientations (s-helo). In: ICIP (2014)

\bibitem{saavedra2017rst}
Saavedra, J.M.: Rst-shelo: sketch-based image retrieval using sketch tokens and
  square root normalization. Multimedia Tools and Applications  \textbf{76}(1),
   931--951 (2017)

\bibitem{saavedra2015sketch}
Saavedra, J.M., Barrios, J.M., Orand, S.: Sketch based image retrieval using
  learned keyshapes (lks). In: BMVC (2015)

\bibitem{schneider2014sketch}
Schneider, R.G., Tuytelaars, T.: Sketch classification and
  classification-driven analysis using fisher vectors. ACM Transactions on
  graphics (TOG)  \textbf{33}(6) (2014)

\bibitem{seff2021vitruvion}
Seff, A., Zhou, W., Richardson, N., Adams, R.P.: Vitruvion: A generative model
  of parametric cad sketches. In: International Conference on Learning
  Representations (2021)

\bibitem{segall1966influence}
Segall, M.H., Campbell, D.T., Herskovits, M.J.: The influence of culture on
  visual perception. Bobbs-Merrill (1966)

\bibitem{smirnov2019deep}
Smirnov, D., Bessmeltsev, M., Solomon, J.: Deep sketch-based modeling of
  man-made shapes. In: ICLR (2021)

\bibitem{song2017spatsematt}
Song, J., Yu, Q., Song, Y.Z., Xiang, T., Hospedales, T.M.: Deep
  spatial-semantic attention for fine-grained sketch-based image retrieval. In:
  ICCV (2017)

\bibitem{SzegedyVISW16}
Szegedy, C., Vanhoucke, V., Ioffe, S., Shlens, J., Wojna, Z.: Rethinking the
  inception architecture for computer vision. In: CVPR (2016)

\bibitem{taubman2012jpeg2000}
Taubman, D., Marcellin, M.: JPEG2000 image compression fundamentals, standards
  and practice: image compression fundamentals, standards and practice,
  vol.~642. Springer Science \& Business Media (2012)

\bibitem{tolias2017asymm}
Tolias, G., Chum, O.: Asymmetric feature maps with application to sketch based
  retrieval. In: CVPR (2017)

\bibitem{vaswani2017attention}
Vaswani, A., Shazeer, N., Parmar, N., Uszkoreit, J., Jones, L., Gomez, A.N.,
  Kaiser, {\L}., Polosukhin, I.: Attention is all you need. NeurIPS  (2017)

\bibitem{wallace1992jpeg}
Wallace, G.K.: The jpeg still picture compression standard. IEEE transactions
  on consumer electronics  \textbf{38}(1) (1992)

\bibitem{wu2015offline}
Wu, J., Wang, C., Zhang, L., Rui, Y.: Offline sketch parsing via shapeness
  estimation. In: AAAI (2015)

\bibitem{shapewords}
Xiao, C., Wang, C., Zhang, L., Zhang, L.: Sketch-based image retrieval via
  shape words. In: ACM ICMR (2015)

\bibitem{xu2021multitrans}
Xu, P., Joshi, C.K., Bresson, X.: Multigraph transformer for free-hand sketch
  recognition. IEEE TNNLS  (2021)

\bibitem{yang2021sketchaa}
Yang, L., Pang, K., Zhang, H., Song, Y.Z.: Sketchaa: Abstract representation
  for abstract sketches. In: ICCV (2021)

\bibitem{Yu2017}
Yu, Q., Song, Y.Z., Xiang, T., Hospedales, T.M.: Sketchx! - shoe/chair
  fine-grained sbir dataset (2017)

\bibitem{yu2017sketchanet}
Yu, Q., Yang, Y., Liu, F., Song, Y.Z., Xiang, T., Hospedales, T.M.:
  Sketch-a-net: A deep neural network that beats humans. IJCV  (2017)

\bibitem{zeng2020bundle}
Zeng, H., Joseph, K., Vest, A., Furukawa, Y.: Bundle pooling for polygonal
  architecture segmentation problem. In: CVPR (2020)

\end{thebibliography}

\appendix

\title{Abstracting Sketches through Simple Primitives\\-\\\large{Supplementary Material}}
\titlerunning{Abstracting Sketches through Simple Primitives}

\author{}
\institute{}
\authorrunning{S. Alaniz et al.}

\maketitle

\section{Network Architecture Details}
For all our networks that take sketches as input, we use the Transformer~\cite{vaswani2017attention} architecture with GeLU activations~\cite{hendrycks2016gelu}, 8 self-attention heads and a layer size of 128 for the self-attention layers and 512 for the fully-connected layers. Our network $f$ has 6 Transformer layers and embeds strokes by adding a single embedding token to the input sequence of points. We find that adding positional embeddings to encode the order of points does not improve the performance, so we only use the coordinates as input for each point.

We train a sketch classifier on Quickdraw with the same architecture as our stroke encoder, i.e. we use the 6-layer Transformer architecture. The first three layers embed strokes and the last three layers 
embed the full sketch by taking the sequence of stoke embeddings as input.
First, we pass the sequence of points of individual strokes (or primitives in PMN) through the first three Transformer layers without positional embedding. To integrate stroke relative positions, the global position and scale are mapped linearly and added to the stroke embeddings. We then pass this stroke embedding sequence to the last three Transformer layers.
A linear layer maps the final sketch embedding to the class logits before taking the cross entropy loss on the ground truth class label.
When partial sketches are evaluated, we mask out unused strokes or sub-strokes at the input to obtain the predicted class logits.

For FG-SBIR, we train a Siamese network~\cite{song2017spatsematt} on the original training sketches with the same architecture of \cite{MuhammadYSXH18,BhuniaYHXS20}, based on the InceptionV3~\cite{SzegedyVISW16} architecture with an embedding size of 128. Since the network acts on natural images and images of sketches, we render our primitive reconstructions to images when evaluating them on the task. Similarly, partial sketches are rendered and fed to the CNN to obtain the retrieval scores for different budget and when applying DSA and GDSA.

\section{Compatibility function}
Our proposed PMN model uses a compatibility function $\phi$ to choose which primitive to match to a human stroke. In theory, the model can also be trained without $\phi$, where the loss functions is applied to each transformed primitive independently, regardless of how well a primitive fits a human stroke. At inference time, without $\phi$ the distance transform needs to be calculated for each transformed primitive to determine which one best fits the human stroke.

\begin{table}[t]
    \centering
    \begin{tabular*}{.7\columnwidth}{@{\extracolsep{\stretch{0.1}}}*{1}{r}@{}@{\extracolsep{\stretch{0.1}}}*{3}{l}@{}}
        Batch size & with $\phi$ & without $\phi$ & factor\\
        \midrule
        32 & 13.14 ms & 118.82 ms & 9.04$\times$\\
        64 & 14.28 ms & 226.46 ms & 15.85$\times$ \\
        128 & 20.16 ms & 449.22 ms & 22.28$\times$ \\
        256 & 34.79 ms & 905.66 ms & 26.03$\times$\\
        512 & 65.23 ms & out-of-memory & -\\[2mm]
    \end{tabular*}%
    \caption{Average time is milliseconds for a forward pass of PMN using the compatibility function (with $\phi$) or not using it (without $\phi$) on a V100 GPU.}
    \label{tab:phi}
\end{table}

Using $\phi$ brings two main advantages. The first is reducing the inference time, without requiring the computation of the distance transform, and the second is providing a better definition of the training loss.  
Firstly, at inference time, we do not require the distance transform to be computed which is the most computationally expensive operation in our pipeline. To quantify this speed-up obtained by using $\phi$, we measured the time a forward pass takes on a V100 GPU (in milliseconds) with and without $\phi$ in Table~\ref{tab:phi}. We see a speed-up of an order of magnitude at small batch sizes of 32 to up to 26 times faster inference time at a batch size of 256. Using a batch size of 512 is not possible without $\phi$ as the 32GB of memory of the V100 is not sufficient to calculate all required distance transforms. On the other hand, we can use a batch size of up to 16384 when using $\phi$ (tested at powers of two).

Secondly, without $\phi$, the loss cannot reach zero, due to the distance transform between target strokes and their most different primitives (e.g. a circle-like human stroke vs the "line" primitive). With $\phi$ instead, the loss can become close to zero since only the most compatible primitives will be used to compute the loss. While we found no clear difference between the two strategies in terms of overall performance on downstream tasks, with $\phi$ the training loss becomes more expressive when comparing varying configurations and it avoids eventual loss spikes caused by matching primitives to strokes of very different shape.

\section{Affine Transformation}

The affine transformation applied on primitives to reconstruct human strokes differs when computing the loss and when recreating a whole sketch. During training, human strokes and primitives are normalized to the range [-1, 1] while retaining the their aspect ratio by subtracting the mean of its points $\mu$ and then dividing by the size of the longest side $w$. The function $h(z^h_p, z^h_s)$ predicts the transformation $T^{p}_{s}$ to align $p$ with $s$ on this normalized scale.
When reconstructing full sketches, primitives are first transformed by $T^{p}_{s}$ followed by denormalizing based on the mean and size of the human stroke to obtain the transformed primitive $pT^{p}_{s}*w_s+\mu_s$. In practice, we combine the scaling factor $w_s$ and the translations $\mu_s$ into the transformations matrix $T^{p}_{s}$ before applying it to $p$ as it also assures that we always use at most six floating point values (maximum number of parameters of a 2D transformation matrix) for our fixed budget communication messages.

The affine transformation predicted by $h$ can be defined in several different ways. We do not allow arbitrary affine transformations in order to retain similarity with the original shape. For instance, if the scaling factors are not controlled, any shape can be collapsed into a line by applying a small factor on one of the axis. Therefore, we restrict the scaling transformation to be a proportional scale where one axis is scaled by a value between 0.05 and 1 while the other is fixed (at 1).
Since scaling alone does not provide enough flexibility to fit primitives to strokes, we experiment with combining scaling with rotation and shear transformations. As reported in Table~\ref{tab:transforms}, the composite transformation of rotate-scale-rotate works best and is chosen for all of our experiments. Notably, these transformations are applied in order, but except for rotate-scale-rotate, changing the order of the transformations, does not have a significant impact on the performance.

\begin{table}[t]
    \centering
    \begin{tabular*}{.8\columnwidth}{@{\extracolsep{\stretch{0.1}}}*{4}{l}@{}}
    \multirow{2}{*}{Transformation} & 
    \multicolumn{3}{c}{Budget}\\
         & 10\% & 20\%  & 30\%\\
        \midrule
        rotate & 44.05 & 64.35 & 73.12 \\
        scale & 57.39 & 69.59 & 73.92 \\
        shear & 57.38 & 74.52 & 80.93\\
        scale, rotate & 64.75  & 80.08 & 85.46 \\
        shear, rotate & 64.69 & 81.74 & 87.86 \\
        shear, scale & 65.99 & 82.12 & 87.69\\
        shear, scale, rotate & \textbf{67.11}  & \textbf{83.73} & 88.86 \\
        rotate, scale, rotate & 67.08  & 83.69 & \textbf{89.15} \\[2mm]
    \end{tabular*}%
    \caption{Classification accuracy on Quickdraw at budgets of 10\%, 20\% and 30\% for different types of transformations learned by $h$.}
    \label{tab:transforms}
\end{table}

\section{Additional Quickdraw Results}

\begin{figure}[ht]
    \centering
    \vspace{-2mm}
    \includegraphics[width=.6\linewidth]{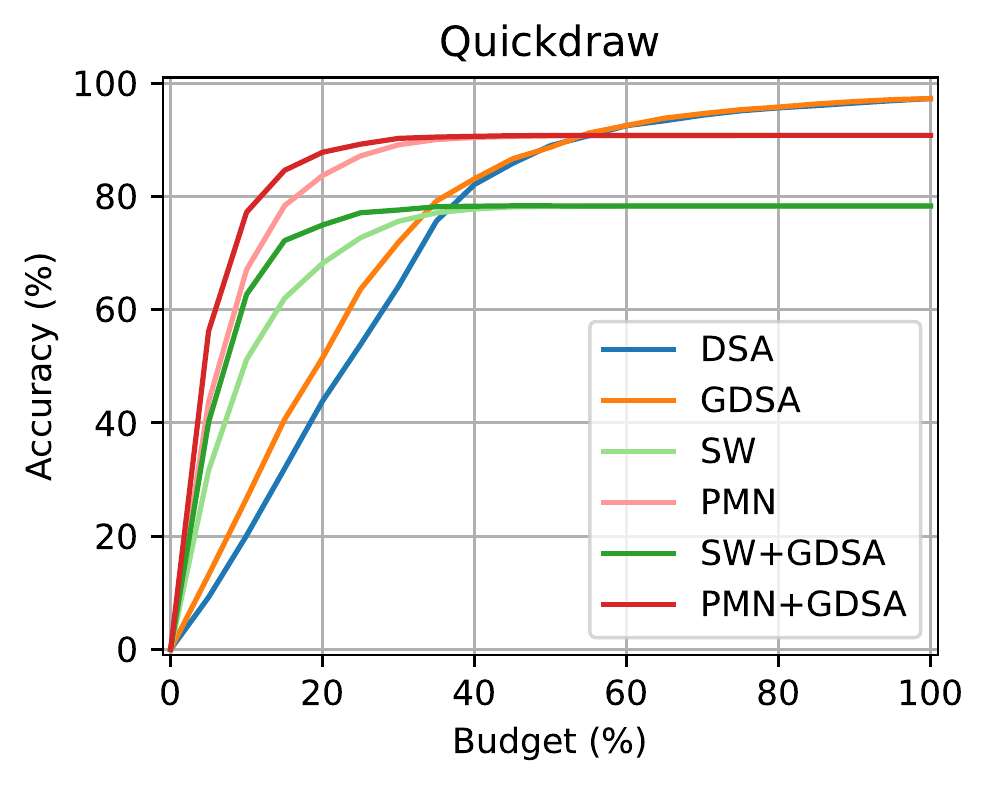}
    \vspace{-4mm}
    \caption{Classification accuracy on Quickdraw at varying budgets between 0\% and 100\% evaluated with a classifier trained on the original human-drawn sketches.}
    \label{fig:quickdraw_budget}
\end{figure}

\textbf{All budget levels.} Figure~\ref{fig:quickdraw_budget} shows the performance of all evaluated methods at different budget levels between 0\% and 100\%. It illustrates the difference between selection-based and shape-based methods. While selection-based methods steadily increase in classification accuracy as the budget increases, shape-based methods have a more steep increase in the beginning and flatten off afterwards, making them favorable in low-budget regimes. As PMN performs lossy compression of the sketches, it requires at most a budget of 70\% to abstract the whole sketch. The intersection of PMN with GDSA is at around 55\% budget.

\noindent{\textbf{Quickdraw-345.} In the main paper, we follow~\cite{MuhammadYSXH18} and we use Quickdraw with 9 classes. Here, we also train and evaluate our PMN model and the compared methods on the Quickdraw dataset with 345 classes. Table~\ref{tab:cls_acc_qd345} shows the results. The trend is consistent with Quickdraw-9, with PMN+GDSA performing the best, and shape-based abstraction outperforming selection-based strategies at low budgets.}

\begin{table}[ht]
    \centering
    \begin{tabular*}{.8\columnwidth}{@{\extracolsep{\stretch{0.1}}}*{6}{c}@{}}
    \multicolumn{2}{l}{Abstraction method} & 
    \multicolumn{4}{c}{Budget (\%)}\\
        Type & Name & 10 & 20 & 30 & 100\\
        \midrule
        \multirow{2}{*}{\shortstack[l]{Selection}} 
         & DSA~\cite{MuhammadYSXH18} & 1.18 & 2.78 & 7.22 & \multirow{2}{*}{\textbf{70.12}}\\
         & GDSA~\cite{MuhammadYHXS19} &  1.39 & 3.45 & 9.04 & \\ [1mm]
        \multirow{2}{*}{\shortstack[l]{Shape}} &SW~\cite{shapewords} & 5.56 & 13.51 & 18.95 & 23.17\\
         & PMN &  11.45  & 25.50 & 33.33 & 38.55 \\ [1mm]
        \multirow{2}{*}{\shortstack[l]{Selection\\+Shape}} &SW+GDSA & 5.92 & 14.82 & 20.12 & 23.17\\
         & PMN+GDSA& \textbf{13.43} & \textbf{27.80} & \textbf{34.87} & 38.55 \\ \hline
    \end{tabular*}\vspace{5pt}
    \caption{Classification accuracy on the Quickdraw345 dataset at budgets of 10\%, 20\% and 30\% evaluated with a classifier trained on the original human-drawn sketches.}
    \vspace{-20pt}
    \label{tab:cls_acc_qd345}
\end{table}

\section{Additional FG-SBIR Results}

In the main paper we report the results of sketch-based image retrieval on top-10 accuracy, following previous works \cite{MuhammadYSXH18}. For completeness, Table \ref{tab:sbir-top1} shows the results for top-1 retrieval accuracy. With this metric, we observe the same trend in all datasets, with PMN+GDSA achieving the best results at low budgets.

{Additionally, apart from the \textit{Selection}-based and \textit{Shape}-based abstraction methods discussed in the main paper, \textit{On-the-Fly Fine-Grained Sketch Based Image Retrieval} (OTF)~\cite{BhuniaYHXS20} proposes a \textit{Finetuning}-based approach that can be employed specifically for the FG-SBIR. Such a method does not learn to abstract, but finetunes the embedding network with partial sketches, optimizing the FG-SBIR ranking to better retrieve their respective images.}

Results for OTF are added to Table~\ref{tab:sbir-top1}. Since shape-based abstraction is orthogonal to finetuning, we also evaluate the combination of SW/PMN with OTF for FG-SBIR.
OTF generally performs well when there is a shift in the data distribution fed through the sketch embedding network. For instance, at 10\% budget on ShoeV2 and ChairV2, OTF outperforms GDSA as finetuning the embedding works better than selecting more relevant parts of the sketch. This gap closes as we increase the budget, and at 30\% GDSA already performs better than OTF on both datasets. Similarly, OTF boosts SW more than our PMN when combined, as the sketches reproduced by SW less accurately resemble the original sketches while the reconstructions of our model stay closer to the original data distribution.

\begin{table}[ht]
     \centering
    \setlength{\tabcolsep}{4pt}
    \renewcommand{\arraystretch}{1}
    \resizebox{\linewidth}{!}{
    \begin{tabular}{l l c c c c c c c c c}
    & & \multicolumn{4}{c}{ShoeV2, Budget (\%)}&& 
    \multicolumn{4}{c}{ChairV2, Budget (\%)}\\\cmidrule{3-6}\cmidrule{8-11}
    Type & Name & 10 & 20  & 30 & 100 &&10 & 20  & 30 & 100\\\cmidrule{1-6}\cmidrule{7-11}
        Finetuning & OTF & 2.40 & 3.45 & 5.11 & \textbf{36.49} && 3.38 & 5.56 & 8.98 & \textbf{53.56}\\ [1mm]
        \multirow{2}{*}{{Selection}} & DSA\cite{MuhammadYSXH18} & 1.35 & 2.40 & 4.05 & \multirow{2}{*}{\textbf{36.49}} && 2.48  & 6.19 & 10.22 & \multirow{2}{*}{\textbf{53.56}} \\
         & GDSA \cite{MuhammadYHXS19}& 1.86 & 3.45 & 6.46 & && 2.79 & 7.43 & 12.07 &\\ [1mm]
        \multirow{2}{*}{{Shape}} & SW & 3.30 & 6.16 & 7.96 & 9.11 && 8.98 & 14.24 & 16.72 & 17.85\\
        &PMN & 6.76 & 16.07 & 18.17 & 20.04 && 16.41 & 31.89 & 35.91 & 37.53\\ [1mm]
        Shape& SW+OTF & 4.80 & 8.41 & 10.66 & 11.92 && 9.91 & 17.03 & 18.89 & 20.31 \\
        +Finetuning&PMN+OTF & 9.16 & 17.17 & 18.92 & 20.77 && 16.72 & 31.89 & 35.15 & 38.04\\ [1mm]
        Shape& SW+GDSA & 4.20 & 7.21 & 8.56 & 9.11 && 9.29 & 14.86 & 17.03 & 17.85 \\
        +Selection&PMN+GDSA & \textbf{9.61} & \textbf{17.37} & \textbf{19.22} & 20.0 && \textbf{20.74} & \textbf{33.75} & \textbf{36.84} & 37.53 \\
        \hline
    \end{tabular}%
    }\vspace{5pt}
    \caption{Top-1 accuracy for FG-SBIR on ShoeV2 and ChairV2.}    \label{tab:sbir-top1}
\end{table}

\end{document}